\pdfoutput=1 
\documentclass[journal,twoside,web]{IEEEtran}
\usepackage{cite}
\usepackage{amsmath,amssymb,amsfonts}
\usepackage{algorithmic}
\usepackage{graphicx}
\usepackage{textcomp}

\newtheorem{definition}{Definition}

\newtheorem{remark}{Remark}
\begin{document}

\title{Rethinking Saliency Map: An Context-aware Perturbation Method to Explain EEG-based Deep Learning Model}
\author{Hanqi Wang, Xiaoguang Zhu, Tao Chen, Chengfang Li, and Liang Song
\thanks{This paragraph of the first footnote will contain the date on 
which you submitted your paper for review. It will also contain support 
information, including sponsor and financial support acknowledgment. For 
example, ``This work was supported in part by the U.S. Department of 
Commerce under Grant BS123456.'' }
\thanks{The next few paragraphs should contain 
the authors' current affiliations, including current address and e-mail. For 
example, F. A. Author is with the National Institute of Standards and 
Technology, Boulder, CO 80305 USA (e-mail: author@boulder.nist.gov). }
\thanks{S. B. Author, Jr., was with Rice University, Houston, TX 77005 USA. He is 
now with the Department of Physics, Colorado State University, Fort Collins, 
CO 80523 USA (e-mail: author@lamar.colostate.edu).}
\thanks{T. C. Author is with 
the Electrical Engineering Department, University of Colorado, Boulder, CO 
80309 USA, on leave from the National Research Institute for Metals, 
Tsukuba, Japan (e-mail: author@nrim.go.jp).}}

\maketitle

\begin{abstract}
Deep learning is widely used to decode the electroencephalogram (EEG) signal. However, there are few attempts to specifically investigate how to explain the EEG-based deep learning models. We conduct a review to summarize the existing works explaining the EEG-based deep learning model. Unfortunately, we find that there is no appropriate method to explain them. Based on the characteristic of EEG data, we suggest a context-aware perturbation method to generate a saliency map from the perspective of the raw EEG signal. Moreover, we also justify that the context information can be used to suppress the artifacts in the EEG-based deep learning model. In practice, some users might want a simple version of the explanation, which only indicates a few features as salient points. To this end, we propose an optional area limitation strategy to restrict the highlighted region. To validate our idea and make a comparison with the other methods, we select three representative EEG-based models to implement experiments on the emotional EEG dataset DEAP. The results of the experiments support the advantages of our method.
\end{abstract}

\begin{IEEEkeywords}
Explaining deep learning, EEG-based deep learning model, Perturbation-based method, Saliency map
\end{IEEEkeywords}

\section{Introduction}
\label{sec:introduction}
\IEEEPARstart{E}{lectroencephalogram} (EEG) is an easily accessible measure of the electrical fields in the active brain \cite{r1}. It has been widely used to decode the brain activities in many applications, e.g., emotion recognition \cite{r26}. EEG data is characterized by the low signal-to-noise ratio (SNR) \cite{r29,r30} and high dimensionality \cite{r1,r31}, which reduces the performance of the traditional technologies. For this reason, many researchers attempt to introduce deep learning into this area, because deep learning shows robust capacity of learning from complex data end-to-end. In many applications, the deep learning based algorithm outperforms state-of-the-art results \cite{r15,r16,r17,r18,r19,r20}. Their performance proves the success of the EEG-based deep learning model. However, the explanation for the EEG-based deep learning model is not fully studied yet, limiting the transparency of the proposed model. Moreover, a clear explanation revealing how the deep learning model works can also help us predict the underlying rules of cognitive process or find the characteristics of brain activities. With the concern to these points, it is necessary to find how we can explain the deep learning model for EEG analysis. 

\begin{figure}[!t]
\centerline{\includegraphics[width=\columnwidth]{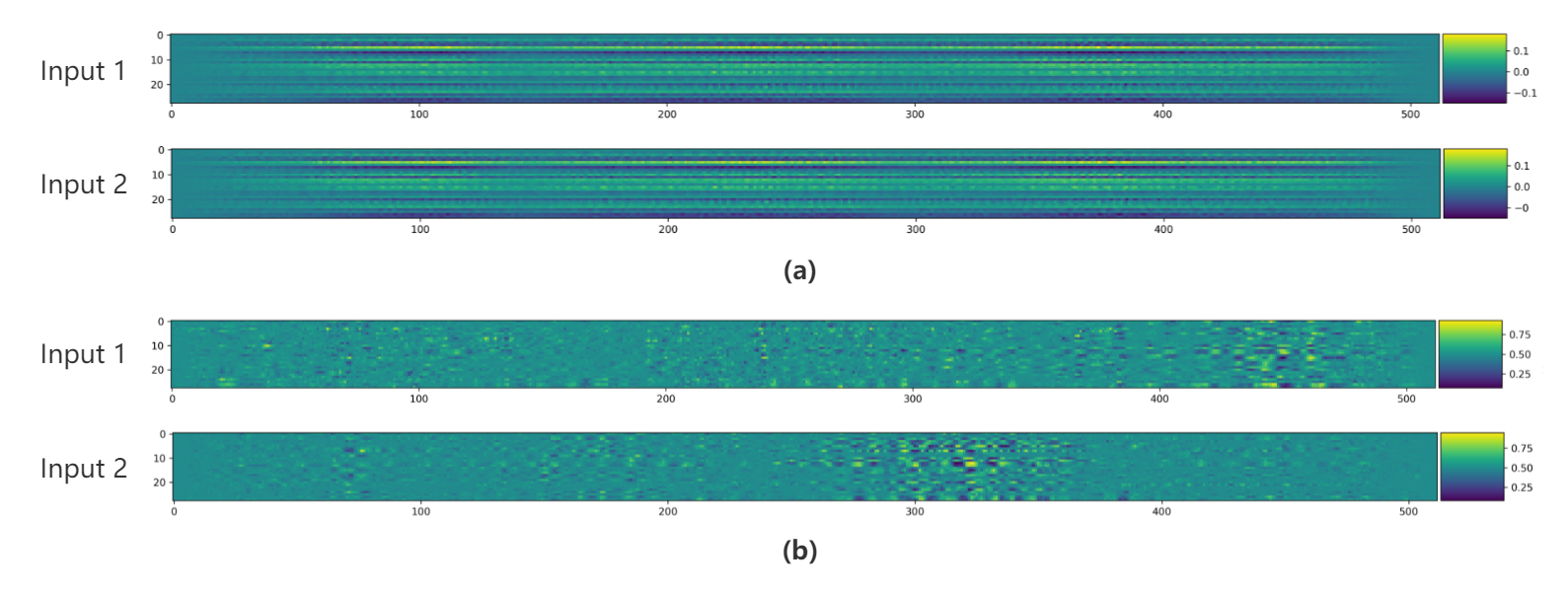}}
\caption{The comparison of generated saliency maps using our method and the gradient-based method on the TSception model and DEAP dataset. Panel (a) demonstrates the result of the gradient-based method, and panel (b) demonstrates the result of our method. In this comparison, we select two input samples at random. And then, we generate saliency maps for these two samples, using the gradient-based method and our method, respectively. We can see that the gradient-based method generates two very similar saliency maps for the various two input samples, proving that the gradient-based method cannot identify the saliency for individual input well.}
\label{fig1}
\end{figure}

An explanation for a deep learning model can be formulated as finding how a deep learning model responds to the input data \cite{r6,r7,r32}. For this purpose, the existing works usually resort to generating a saliency map. Here, the saliency can be understood as the relevance of feature to the output of the model, and the saliency map highlights the region relevant to the output of the model. Most of the current works identify the saliency map through gradient-based method or perturbation-based method. Most of the previous works~\cite{r2,r5,r8,r9} aiming at explaining EEG-based deep learning models adopt the gradient-based method. The gradient-based method calculates the gradient of output with respect to the input, using it as an indicator of saliency \cite{r21}. Because the large gradient represents rapid local variation, the region with a large gradient is considered to have a larger impact on the output. However, the gradient-based method captures the average property of the network, not the instance-level saliency \cite{r6}. This weakness can be understood in a simple example. Considering the model $f$ we want to explain is a linear model, e.g., multilayer perceptron (MLP), the generation of output can described as: $f(x_{0})=<w,x_{0}>+b$. The $x_{0}$ is input, the $w$ is the parameter of the model, and $b$ is a constant scalar. The gradient of output with respect to input is : $\nabla f(x_{0})=w$. We can see that the gradient-based saliency map cannot represent how the linear model responds to a specific input $x_{0}$. Although the activation function embedded in the deep learning model alleviates this effect, this problem is not eliminated. In Fig.\ref{fig1}, we show an example for this problem. The previous work \cite{r6} has reported that sometimes the gradient-based saliency map responds strongly to some regions without important information as well. Unfortunately, this shortcoming of the gradient-based method is particularly intolerable when we explain EEG-based deep learning model. Due to the non-stationary nature of EEG signals \cite{r27,r28}, the characteristics of EEG signals vary rapidly from time step to time step, subject to subject. Consequently, a saliency map that fails to demonstrate the instance-level responding mechanism of the model cannot be considered as a reliable explanation.

The perturbation-based method is a currently popular approach to explaining the deep learning model. This framework uses the effect of perturbation in the input on the output as an indicator of saliency \cite{r22}. A region in the input is assumed as salient if applied perturbation in this region causes a great effect on the output. As far as we know, there is no attempt to investigate saliency map using the perturbation-based method for EEG-based models. But some works \cite{r3,r4,r10} adopt the key assumption of perturbation-based method to detect local saliency in frequency domain. In their framework, they apply random noise or replacement to some individual frequency band, respectively, to investigate the impact of each band on the output. Their works make an interesting effort to propose an explanation method for the EEG-based deep learning model. However, their local perturbation method cannot identify saliency from a global perspective. Furthermore, their perturbation method is a relatively rough approach. It fails to provide an accurate and reasonable explanation and risks triggering artifacts in the deep learning model. Considering these weaknesses, we cannot regard their works as a promising way to investigate saliency problem for the EEG-based model. 

Our review suggests a paucity of a reliable method to identify the saliency for the EEG-based models. Recently, mask perturbation is proposed in \cite{r6,r7,r11} for image data. This modified perturbation-based method introduces a mask to jointly perturb the input and search for the saliency map through backward propagation. This framework improves the existing perturbation-based methods, resulting in a more powerful explanation \cite{r12,r13,r14}. Their works inspire us to investigate this problem from a new perspective. However, their works cannot address the characteristics of EEG data. It motivates us to propose a new method specifically for explaining the EEG-based model.

\begin{figure}[htbp]
\centerline{\includegraphics[width=\columnwidth]{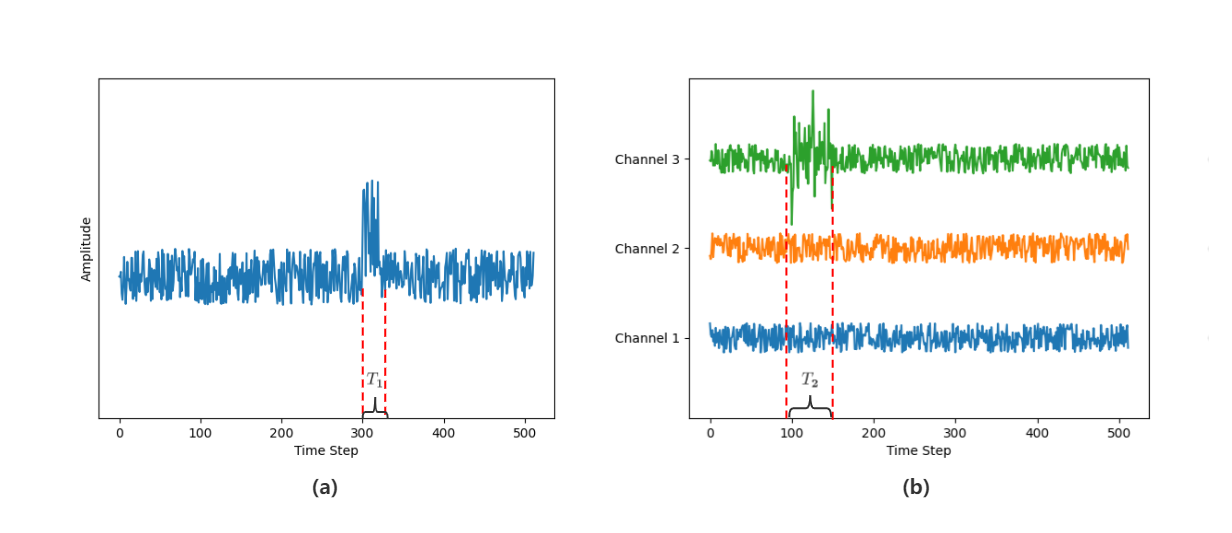}}
\caption{The illustration of the importance of the context in EEG. The part (a) shows an example of the temporal context. And the part (b) shows an example of spatial context. $T_1$ and $T_2$ represent time period. Various channels record the EEG signals collected from various regions on the scalp.}
\label{fig11}
\end{figure}

Compared with image data, EEG data is quite different since EEG data is characterized by temporal dependency and spatial dependency~\cite{r1,r31}. In other words, the surrounding features influences how the model decides the importance of the features. This characteristic of EEG induces a concern for the context, including temporal context and spatial context. In~\ref{fig11}, we illustrate this problem with two simplified examples. To begin with, the variation of EEG over time usually contains rich information. As we can see, the EEG signal suddenly rises within the time period $T_1$ in Fig~\ref{fig11} (a). Combined with the low-level amplitude on surrounding time steps, this sudden change might form a special pattern which is crucial in the eyes of the deep learning model. Besides that, the features in EEG are spatially correlated, either. For example, many researchers have reported that spatial asymmetry in EEG characterize the emotional status of the brain~\cite{r33,r39,r40}. We exemplify this observation in Fig~\ref{fig11} (b). As we can see, the EEG signals collected from various region on the scalp show an obvious asymmetry on spatial dimension during time period $T_2$. This asymmetry could be an hint for a particular emotional status. Therefore, the spatial context in EEG have a great implication for different cognitive activities, making it indispensable for understanding the EEG-based deep learning model. In a word, the context information is crucial for the output issued by the EEG-based deep learning model. Hence, a reliable explanation for the EEG-based model is required to embody the dependencies between different features, i.e., the influence of the context.

Furthermore, the context information is also crucial for dealing with artifacts in the deep learning model. The authors in \cite{r6,r7} point out that perturbation operation generating abnormal samples is at risk of triggering the artifacts in the deep learning model. Many works \cite{r23,r24,r25} have found that the affected model will generate an unexpected or nonsensical output. This observation indicates that the explanation is also damaged by the artifacts induced by perturbation. It makes the obtained explanation loses its generality and interpretability. The reason for triggering artifacts can be attributed to the abnormality of the perturbed input because the model is trained on normal samples. Thus, the second purpose of collecting context information is to simulate the characteristic of normal EEG data. As a consequence, the perturbed input will get close to the distribution of normal samples, lowering the risk of triggering artifacts. 

\begin{figure}[htbp]
\centerline{\includegraphics[width=\columnwidth]{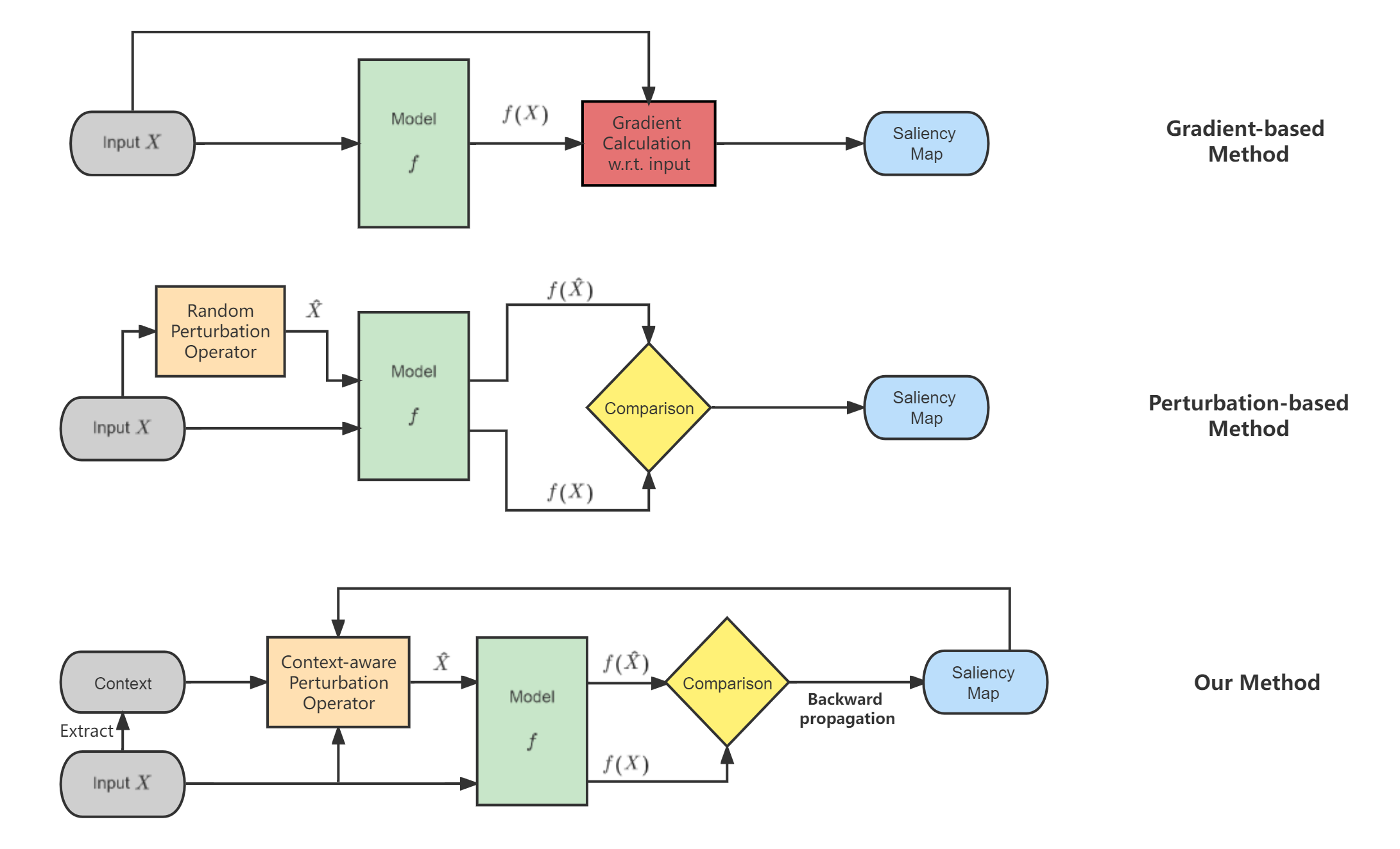}}
\caption{The flowchart of the gradient-based method, traditional perturbation-based method, and our method. $f$ is the target deep learning model, $X$ is the input, and $\hat{X}$ is the perturbed input. And the output of $f$ taking $X$ as input is marked as $f(X)$. The output of $f$ taking $\hat{X}$ as input is marked as $f(\hat{X})$}
\label{fig2}
\end{figure}

For this reason, we propose a context-aware perturbation method to explain the EEG-based deep learning model in this paper. In our framework, the context information is used to build the perturbation for the input. It includes the context in consideration of our method when searching for the estimate. In Section \ref{subsec1}, we make a detailed analysis of the effect of context-aware perturbation. To collect context information, our context-aware perturbation method adopts pooling operations on temporal and spatial dimensions, respectively. And then, our method adopts a soft attention mechanism to adaptively fuse them, making the synthesized information more representative of the local context. In some scenarios, another concern over an explanation for EEG-based deep learning model is the legibility of the output saliency map. A raw EEG signal input usually consists of a large number of feature entries. Sometimes, the number of highlighted features might be huge, resulting in an ambiguous indication. So, the user might want to highlight a small region of the input for a simple version of the explanation. In order to filter the relevant region, we provide an optional area limitation strategy to preserve the highly salient region and encourage the weakly salient region to be turned off. This simplification preserves the reliability of the explanation while reducing its complexity. We conduct extensive experiment to validate our ideas and compare our method with the commonly used gradient-based method. The experiment are implemented on a public emotional EEG dataset DEAP, using three representative model with various scales. In those experiments, we adopt either channel reduction or time steps reduction to cause the impact on the model performance. Then, we use the caused impact to measure the saliency of removed elements. The result of our experiments show that our method can effectively identify those salient features. And the result of our experiments also support the 
reliability of our method compared to the gradient-based method. Our contribution can be summarized as the following:
\begin{enumerate}
\item We carefully review the saliency methods which aim to explain EEG-based method. And we also summarize the shortcomings of the current works.

\item We propose the context-aware perturbation to incorporate temporal dependency and spatial dependency. Moreover, we indicate that this perturbation method also help suppress the artifacts in the deep learning model, resulting in a more reliable explanation.

\item We provide a solution to limit the highlighted area in order to improve the legibility. The most salient features are preserved while the weak features are suppressed.

\item We implement experiments to validate our idea and make comparisons with the gradient-based method. The results show the superiority of our method.
\end{enumerate}
\section{Method}
\begin{figure*}[htbp]
\centerline{\includegraphics[width=\textwidth]{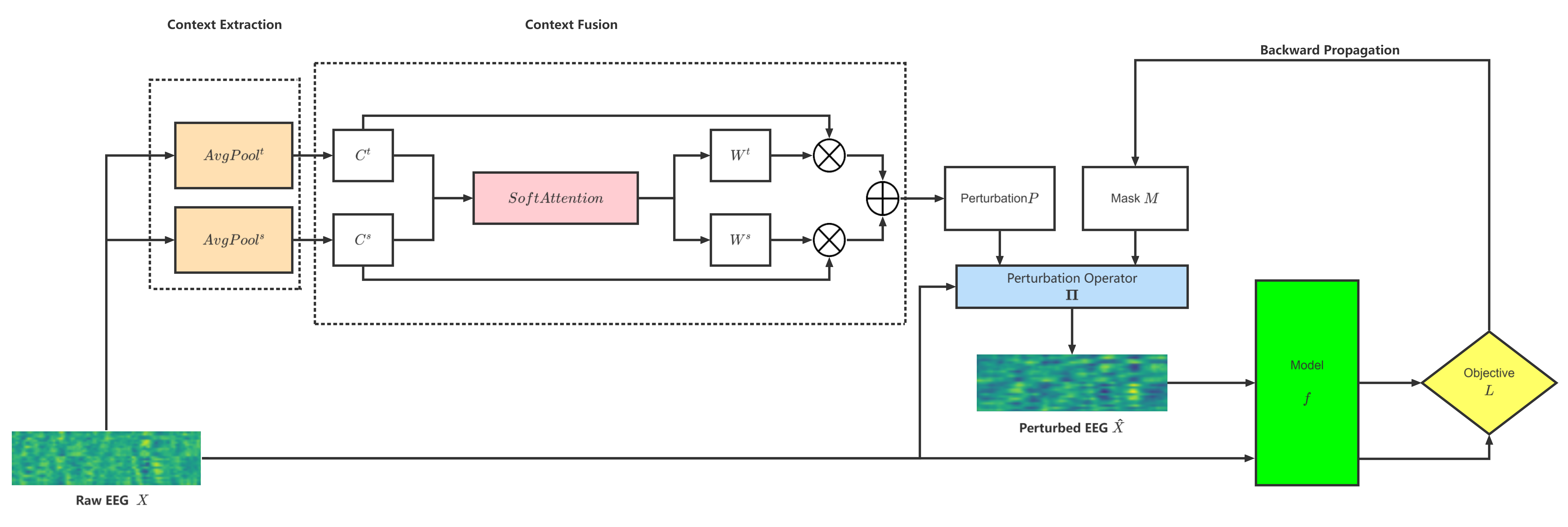}}
\caption{The structure of the proposed context-aware perturbation method. The context extraction module extracts the context information from the raw EEG input $X$ on various dimensions. And then, the context fusion module fuse the extracted context information to obtain the perturbation $P$ using an adaptive strategy. The synthesized perturbation $P$ is brought into the operator $\Pi$. Under the instruction of mask $M$,  The operator $\Pi$ generates the perturbed EEG input $\hat{X}$. Finally, the output of operator $\Pi$ and the raw EEG both enter the black-box model $f$ as input. The difference between their outputs is used to build objective $L$. Following the objective, the backward propagation optimizes the mask $M$ automatically.}
\label{fig10}
\end{figure*}
In this section, we illustrate the details of the proposed context-aware perturbation method. The overall structure can be seen in the Fig~\ref{fig10}. Here, we first introduce the basic concept of proposed context-aware perturbation. And then, we illustrate the context-aware perturbation operation. Following it, we make a analysis to prove the strength of context-aware perturbation. Next, we discuss the optimization strategy of the proposed mask involving the area limitation strategy. 
\subsection{Overview}
In this paper, we use $X$ to denote the input EEG sample, $X\in R^{Ch\times T}$, where $T$ denotes the length of time dimension and $Ch$ denotes the number of channel. It is worth noting that each channel in EEG signal corresponds to the brain activities in a certain area. Thus, the channel dimension can be considered as spatial dimension. we formalize the definition of mask in the following:
\begin{definition}
$M$ is the mask associated with a given input $X$ and a given model $f$, where $M\in [0,1]^{Ch\times T}$. The mask $M$ has the same size as the input. It represents the saliency map of the input. Each element $m_{ch,t}$ in $M$ corresponds to the saliency estimate of the feature $x_{ch,t}$ at time $t$ in channel $ch$. So, the mask can also be regarded as the saliency map we are looking for.
\end{definition}
\begin{remark}
 For a given element $m_{ch,t}$ in $M$, its corresponding feature $x_{ch,t}$ is regraded as salient point if $m_{ch,t}$ is close to 1. Otherwise, the $x_{ch,t}$ would be regraded as the opposite. 
\end{remark}

The mask $M$ will dictate how to perturb the input. The basic assumption for context-aware perturbation is that the model's output is sensitive to the perturbation in the salient region and robust for the perturbation in the non-salient region. Based on this assumption, the strategy of context-aware perturbation operation is to preserve the features whose corresponding mask values are close to 1 during perturbation operation and vice versa. The reason is that this strategy provides access to evaluate the quality of $M$. Suppose the perturbation operation under the guidance of a mask causes a subtle effect on the model's output. In that case, it proves that the mask captures a reliable estimate of the saliency map. That is also the cue to search for the mask in the next step. 

Considering that the effect on the output caused by the perturbation suggests the quality of the mask, we formulate the search for the optimal mask into a machine learning task. Let $\hat{X}$ denotes perturbed input, $f(\hat{X})$ denotes perturbed output, and $f(X)$ denotes original output. We set the $M$ as a learnable parameter matrix and the difference between these two outputs as an objective function. The training objective is to minimize the effect of the perturbation on the output. With this setting, the backward propagation wildly used in machine learning can be introduced to search for optimal $M$ automatically. 
\subsection{Context-aware Perturbation Operation}
Let $\Pi$ denotes the perturbation operator. As shown in (\ref{eq1}), it will act on the input $X$ associated with the $M$ to generate the perturbed input $\hat{X}$. 
\begin{equation}
\hat{X}=\Pi(X,M)
\label{eq1}
\end{equation}
Then, we want to introduce our strategy of utilizing the mask during the perturbation operation. As we mentioned above, The rule of our proposed work is to preserve the salient features indicated by the mask and perturb those non-salient ones. In order to achieve it, our work utilizes the saliency information in the mask as follows:
\begin{equation}
\Pi(X,M)=(1-M)\cdot P+M\cdot X,
\label{eq2}
\end{equation}
where $P\in R^{Ch\times T}$ denotes the perturbation applied on the input. Concretely, for a feature $x_{ch,t}$, its corresponding perturbed feature $\hat{x_{ch,t}}$ is calculate as shown in \eqref{eq3}.
\begin{equation}
\hat{x}_{ch,t}=(1-m_{ch,t})\cdot p_{ch,t}+m_{ch,t}\cdot x_{ch,t}.
\label{eq3}
\end{equation}
As we can see in $(\ref{eq3})$, the perturbation will vanish in location $(ch,t)$ if the $m_{ch,t}=1$. And when the $m_{ch,t}=0$, conversely, the original information in $x_{ch,t}$ will be completely replaced. 

To compute the perturbation $P$, we first compute the context information.
Pooling operation has been proven to be an effective method to extract context information in the deep learning area. Hence, we adopt two 1-D average pooling operations to extract context information on temporal dimension and spatial dimension, respectively, as shown in $(\ref{eq4})$.
\begin{equation}
C^{t}=AvgPool^{t}(X), 
C^{s}=AvgPool^{s}(X),
\label{eq4}
\end{equation}
where $C\in R^{C\times T}$ is the context information, superscript $t$ and $s$ represent temporal dimension and spatial dimension respectively, and $AvgPool$ represents the average pooling operation. Here we keep the shape of each $C$ same as the input $X$. Then, the various context information will be fused to obtain complete context information:
\begin{equation}
P=W^{t}\cdot C^{t}+W^{s}\cdot C^{s},
\label{eq5}
\end{equation}
where superscript denotes the dimension, $W$ denotes the weight matrix for corresponding context information $C$.

Suppose the EEG signal in channel $c$ shows an obvious asymmetry on spatial dimension at time $t$ while the fluctuation of neighboring features is quite mild on the temporal dimension. In that case, it is unfair to treat the various context information equally. In order to capture the context in a more representative way, we need an approach to achieve an adaptive fusion of these two context information. Here, we introduce a soft attention mechanism to compute the final perturbation. For a feature $x_{ch,t}$, the computation for the final perturbation $p_{ch,t}$ in it is shown in $(\ref{eq6})$ and $(\ref{eq7})$.
\begin{equation}
p_{ch,t}=w_{ch,t}^{t}\cdot c^{t}_{ch,t}+w_{ch,t}^{s}\cdot c^{s}_{ch,t},\\
\label{eq6}
\end{equation}
\begin{equation}
w_{ch,t}^{t}= \frac{e^{c^{t}_{ch,t}}}{e^{c^{t}_{ch,t}}+e^{c^{s}_{ch,t}}},
w_{ch,t}^{s}= \frac{e^{c^{s}_{ch,t}}}{e^{c^{t}_{ch,t}}+e^{c^{s}_{ch,t}}},
\label{eq7}
\end{equation}
where $c$ denotes corresponding context information, $w$ denotes the weight for $c$, superscript denotes the dimension, and the subscript denotes the location. 

As can be seen, the perturbation applied on the input integrates the context information. It is a crucial step to make search process concerns the context .

\subsection{Analysis of The Proposed Context-aware Perturbation}
\label{subsec1}
In this subsection, we backtrack the gradient of the objective with respect to the mask. This step aims to understand that, under which concern, the backward propagation optimizes the mask.

Let $L$ be a scalar function, representing the training objective. When it searches for the optimal $M$, backward propagation algorithm calculates the gradient of $L$ with respect to learnable parameters $M$. The gradient matrix can be written as:
\begin{equation}
\label{eq11}
\nabla_M L=
	\begin{bmatrix}
	 \frac{\partial L}{\partial m_{1,1} } & \frac{\partial L}{\partial m_{1,2} } & 
	 \cdots &\frac{\partial L}{\partial m_{1,T} }\\
	\frac{\partial L}{\partial m_{2,1} } & \frac{\partial L}{\partial m_{2,2} } & 
	 \cdots &\frac{\partial L}{\partial m_{2,T} }\\
	\vdots & \vdots & \ddots & \vdots\\
	\frac{\partial L}{\partial m_{Ch,1} } & \frac{\partial L}{\partial m_{Ch,2} } & 
	 \cdots &\frac{\partial L}{\partial m_{Ch,T} }
	 \end{bmatrix}
\end{equation}

According to the chain rule, each element $\frac{\partial L}{\partial m_{ch,t}}$ in the gradient matrix can be written as:

\begin{equation}
\label{eq12}
\begin{split}
    \frac{\partial L}{\partial m_{ch,t}}=\sum_{k=1}^{Ch}\sum_{l=1}^T \frac{\partial L}{\partial \hat{x}_{k,l}}\cdot \frac{\partial \hat{x}_{k,l}}{\partial m_{ch,t}}.
\end{split}
\end{equation}
Then ,we substitute $\hat{x}_{k,l}$ in \eqref{eq12} with the \eqref{eq3}. Obviously, $\frac{\partial \hat{x}_{k,l}}{\partial m_{ch,t}}=0$, if $k\neq ch$ or $l\neq t$. Finally, we can get:
\begin{equation}
\label{eq13}
\frac{\partial L}{\partial m_{ch,t}}=\frac{\partial L}{\partial \hat{x}_{ch,t}}\cdot (-p_{ch,t}+x_{ch,t}).
\end{equation}

As can be seen in the \eqref{eq13}, the backward propagation optimize the mask considering the input and the perturbation. Remembering that our context-aware perturbation integrate the context information, we can say that our method is able to perceive the influence of the context. And compared with the gradient-based method, the \eqref{eq13} also demonstrates that our method can concern the individual input.
\subsection{Optimization and Area limitation Strategy}
As we stated before, the difference between the $f(x)$ and $f(\hat{x})$ will be used to design the error loss $L_{error}$. The concrete design will be slightly different in light of the task. For example, for a classification task, we can use the cross-entropy as objective function:
\begin{equation}
L_{error}=- f(X)\cdot \log f(\hat{X}).
\label{eq8}
\end{equation}

And for a regression task, the objective function can also be replaced by other ones, e.g., mean squared error (MSE).

In some scenarios, the users might want a clear identification of the salient region, For the purpose, we propose an optional area limitation strategy. The use of this term can be decided by the users according to their needs. This strategy aims to encourage most elements in the mask to be 0, only preserving the elements with higher values. In this way, we can identify as few features as possible, finding a clear explanation of the investigated model. The previous works \cite{r14,r24} introduce a binary regularization term to reach this goal. However, in practice, we find that this regularization term harms the effectiveness of the saliency map. Because it sometimes conflicts to minimize the $L_{error}$. We reform the regularization term under this circumstance, making it only suppress the non-salient element. The modified regularization term is shown in $(\ref{eq9})$.
\begin{equation}
L_{area}=\left\|vecsort(M)-r_a\right\|_2,
\label{eq9}
\end{equation}
where $\left\|.\right\|_{2}$ denotes the Euclidean norm. The $vecsort(M)$ means a function reshapes the $M\in R^{Ch\cdot T}$ into a $Ch\times T$ long 1-D vector, and then sort it in a ascending order. The $r_a$ is a $Ch\times T$ long 1-D vector, too. The first $a\times Ch\times T$ elements of $r_a$ are 0, and the last $(1-a)\times Ch\times T$ elements are equal to the corresponding elements in $vecsort(M)$. The hyperparameter $a\in [0,1]$ controls the ratio of elements preserved in the mask. Therefore, the objective function including area limitation strategy is formulated as $(\ref{eq10})$.
\begin{equation}
L= L_{error} + \lambda\cdot L_{area},
\label{eq10}
\end{equation}
where $\lambda$ is the coefficient which control the importance of the regularization term. However, we find that the the regularization term $L_{area}$ is likely to harm the search for optimal $M$, because the regularization brings strong bias towards specific areas based on the initialization of the mask.
\begin{figure}[!t]
\centerline{\includegraphics[width=\columnwidth]{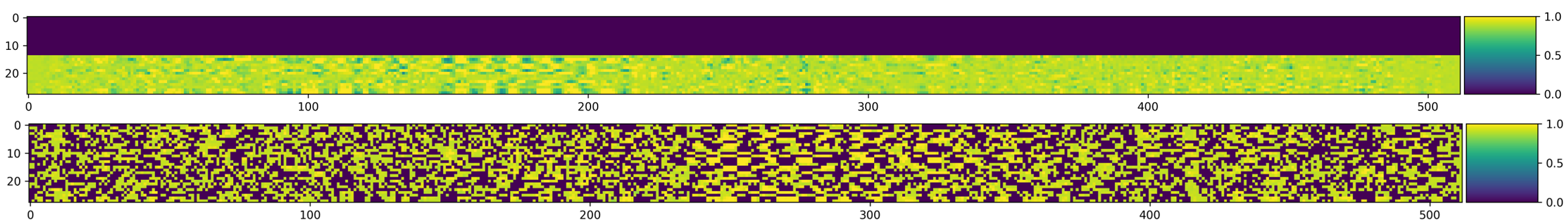}}
\caption{The example of the bias brought by regularization term $L_{area}$. The two saliency maps are generate for the same input using the area limitation strategy. The top one follows the one-stage training in \cite{r14,r24}. And the bottom one follow the proposed two-stage training. As we can see, the term incorrectly suppress the top region of the input, leading to a nonsensical generation. And our solution can avoid such a problem.}
\label{fig3}
\end{figure}

Imaging that we are at the first training epoch, and all the elements in the $M$ are initialized to a fixed value. In that case, the regularization term $L_{area}$ will naively suppress the top part of the input X. This influence will continue to influence the following training process. The unexpected result can be seen in the top of Fig.\ref{fig3}. Though a careful choice for the $\lambda$ can alleviate this problem, this approach can not eliminate it. For this reason, we propose a two-stage training strategy. In the beginning $k$ epochs, we only use the $L_{error}$ as objective to learn a gross estimate of the mask. And then, we shift the objective to $L$ in \eqref{eq10} in the following epochs. In this way, the $L_{area}$ will suppress the non-salient region based on the real estimate instead while considering the error loss $L_{error}$. As shown in bottom of Fig.\ref{fig3}, the two-stage training strategy effectively solves this problem.

\section{Experiment}
\subsection{Dataset and Model Introduction}
we conduct experiment on a public dataset DEAP \cite{r34}, which is commonly used in emotion recognition task. DEAP consists of two dimensions, arousal and valence. The EEG data are collected from 32 subjects. For every subject, the researchers conduct 40 trials in which the subjects are required to report their emotional state after watching the video clips. Each trial has a duration of 60 seconds. And the EEG signals from 32 active electrodes are recorded during the trial. The recorded EEG signals are sampled at a rate of 512Hz. In the experiment, we follow the paradigm reported in \cite{r9,r35} to utilize the data.

\begin{table}[!ht]
    \caption{The number of trainable parameters of the three selected models.}
    \centering
    \begin{tabular}{|c|c|}
    \hline
    Model              & Parameters   \\
    \hline
    EEGNet            & 1210            \\
    \hline
    TSception         & 12563           \\
    \hline
    DeepConvNet       & 281427            \\
    \hline
    \end{tabular}
    
    \label{tab1}
\end{table}

To validate the performance of our method, we carefully select one state-of-the-art deep model targeting at emotion recognition, TSception \cite{r9}, and two commonly used general models, EEGNet \cite{r19} and DeepConvNet \cite{r3}. The model selection is based on two consideration. First, we think our method should be examined on the network designs with various target purposes. Second, we are interested in the effectiveness of our method on the models at different scales. As we can see in Table \ref{tab1}, the three models show an increasing order of magnitude of the trainable parameters. These three models are all working end-to-end. They take the raw EEG data as input and then output scores for each class. All three models are implemented on the DEAP dataset following the same paradigm.

\subsection{Experiment Settings}
We adopt the within-subject classification task to train the investigated models in the experiment. Within-subject classification uses subject-specific data to train the model for each subject. Furthermore, we split each trial into non-overlapping segments, following \cite{r3}. The three models are all trained with the default settings reported in the original papers. In the training process of the mask, we use two pooling operations to extract context information. For pooling on the temporal dimension, the kernel size is 15, and the stride is 1. For pooling on the spatial dimension, the kernel size is 28, and the stride is 1. Then, we use the loss in \eqref{eq8} to measure the error except for the experiment in Section~\ref{area}. For the area limitation strategy, we adopt~\eqref{eq10}, and we set $a=0.5$, $\lambda=0.05$, and $k=100$. 
In the beginning, all the elements in the mask are set to $0.9$. We adopt Adam as an optimizer, with a learning rate equal to 0.1. Moreover, each mask is trained with 200 epochs.
\subsection{Channel Reduction}
\label{Channel Selection}
Many researchers are interested in the spatial characteristic of brain activities. Thus, some researchers \cite{r2} adopt the saliency map to select the most informative channels, and they also adopt channel reduction to validate the effectiveness of their selection. Following their paradigm, we use the saliency map as an indicator to select channels, validating the effectiveness of identified saliency map. The selected channels are removed from the input. The effectiveness of identified saliency map can be seen from the performance of the model using reduced input. Concretely, we first generate an instance-level saliency map for each combination of subject-specific model and each input. Then, we calculate the mean of saliency values along the time dimension on each saliency map $M_{i,j}$, where $i$ denotes the $i_{th}$ subject-specific model and $j$ denotes the $j_{th}$ sample which target at that model. The obtained mean values are regarded as the saliency scores for the corresponding channels. Before we send the samples into the model $f_i$, we implement the removal of channels based on the scores. Following \cite{r2}, we set the features in the removed channel to 0. After the removal, we send the reduced input into the model. Next, we can get the accuracy of the $f_i$ using the reduced input. To give out the overall performance of 32 subject-specific models, we use the mean of the accuracy of each model as the measure of overall performance.
\begin{figure}[!t]
\centerline{\includegraphics[width=\columnwidth]{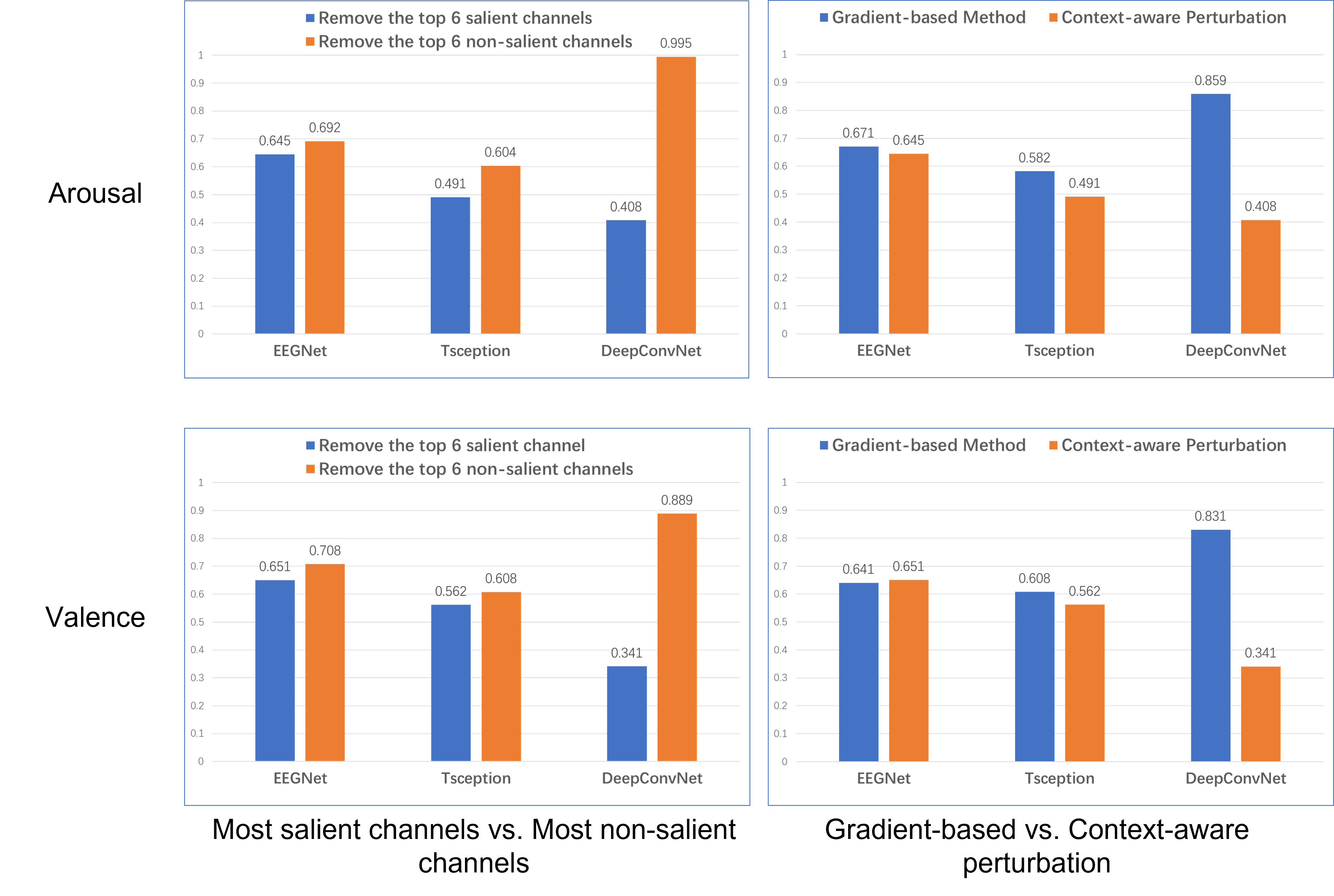}}
\caption{The mean accuracies of model after removal of channels. The rows correspond to the dimensions of the DEAP dataset, and the columns correspond to the two experiments. A lower mean accuracy suggests a more significant impact on the output. According to the definition of saliency, if the removal of some channels leads to a more significant impact on the output, the removed channels are more salient than the others. }
\label{fig4}
\end{figure}

Here, we first implement two experiments. In the first experiment, we remove the most salient six channels and the most non-salient six channels. And in the second experiment, we implement the gradient-based method used in \cite{r2,r5,r8,r9}. We remove the most salient six channels identified by our method and gradient-based method, respectively. The comparisons of both experiments can be seen in Fig \ref{fig4}. The result of the first experiment is shown in the first column. We can see that if we remove the most salient channels, the performance of the models will decrease more than if we remove the most non-salient channels. It means that our proposed method effectively captures the features with different saliency levels. It is an essential requirement for a reliable method of generating saliency map. And the result of the second experiment is shown in the second column. We compare our method with the widely used gradient-based method in the second column. As we can see, the performance of the models will drop more if we remove the channels identified by our method. It means the proposed context-aware perturbation can better identify the salient features than the gradient-based method. Our method performs better with the increase in model scale in both experiments. This observation can be analyzed from the perspective of expressive capacity. The expressive capacity represents the ability of deep learning models to approximate complex problems \cite{r38}. According to \cite{r36,r37}, the parameter scale is usually related to the expressive capacity. So, a model with large-scale trainable parameters is more likely to capture the most informative region in the input. Therefore, the generated saliency map will also concentrate on those informative features. As a result, removing salient channels brings a more severe impact.

\begin{figure}[!t]
\centerline{\includegraphics[width=\columnwidth]{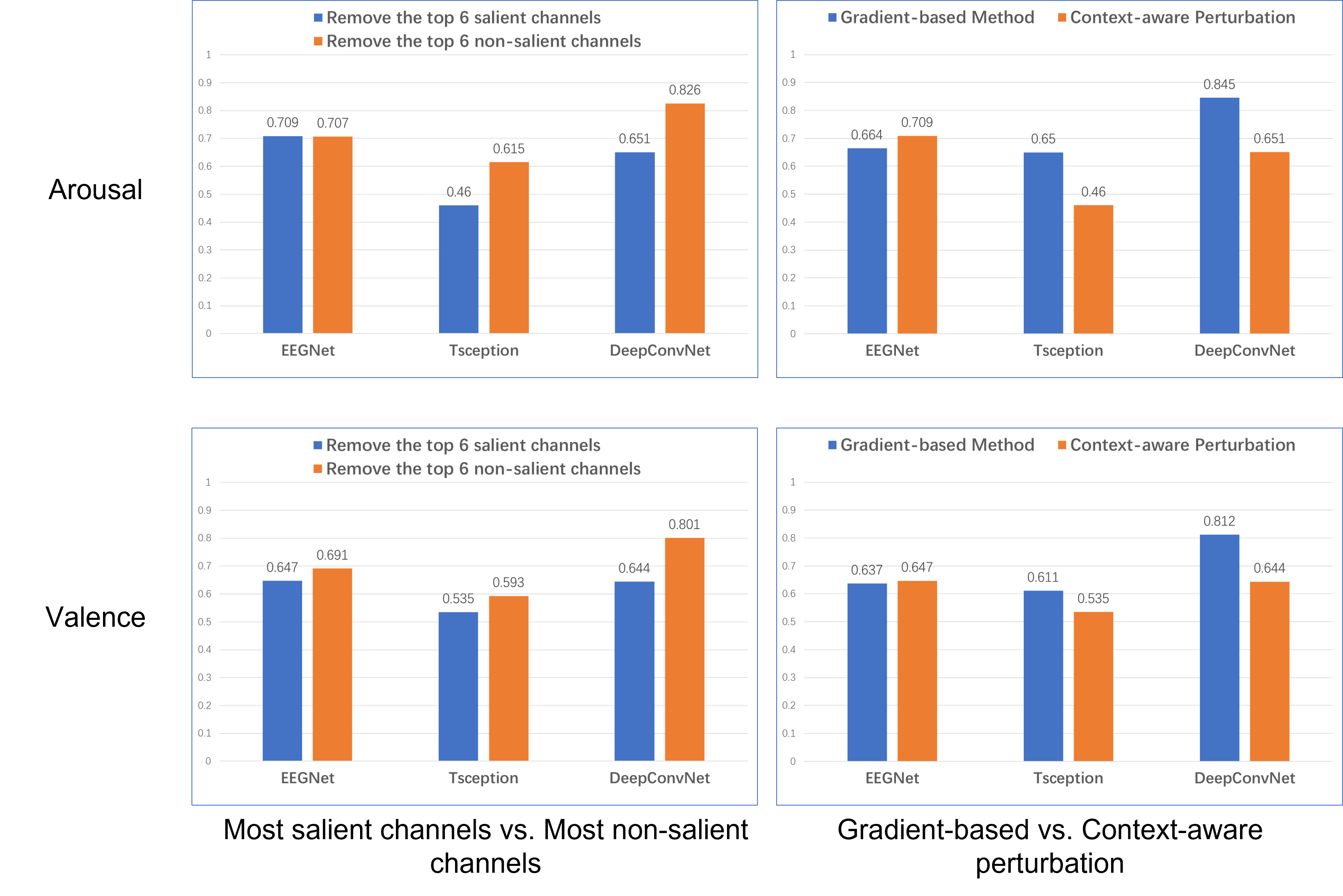}}
\caption{The results of the group-level experiment. In this figure, we show the mean values of accuracy as the result of the group-level experiment. The rows correspond to the dimensions of the DEAP dataset, and the columns correspond to the two experiments.}
\label{fig5}
\end{figure}

Sometimes, a user might want a general indication of salient channels. According to this concern, we generalize the scope of the saliency map from instance-level to group-level. In order to do this, we average the group of masks for the inputs in each subject-specific dataset. And then, we implement channel reduction using the averaged mask, the same as before. Finally, we calculate the mean of 32 accuracy results as in the previous experiment. The results are shown in Fig \ref{fig5}. First, the result in the left column shows that our method is still effective in identifying the salient channels. Then, we can see from the second column that our method outperforms the gradient-based method even at the group level. Although the performance of the gradient-based method is roughly equivalent to our method for EEGNet, the difference is growing more significant with the increase in model scale. Finally, if we make a comparison of the second column in both Fig \ref{fig4} and Fig \ref{fig5}, we will find that the results of the gradient-based method are almost the same. It makes sense only if the gradient-based method generates a very similar saliency map for each input, causing the averaged saliency map to have no significant difference with each instance-level saliency map. This observation further supports the shortcomings of the gradient-based method we mention in Section \ref{sec:introduction}.
\subsection{Time Step Reduction}
\begin{figure}[!h]
\centerline{\includegraphics[width=\columnwidth]{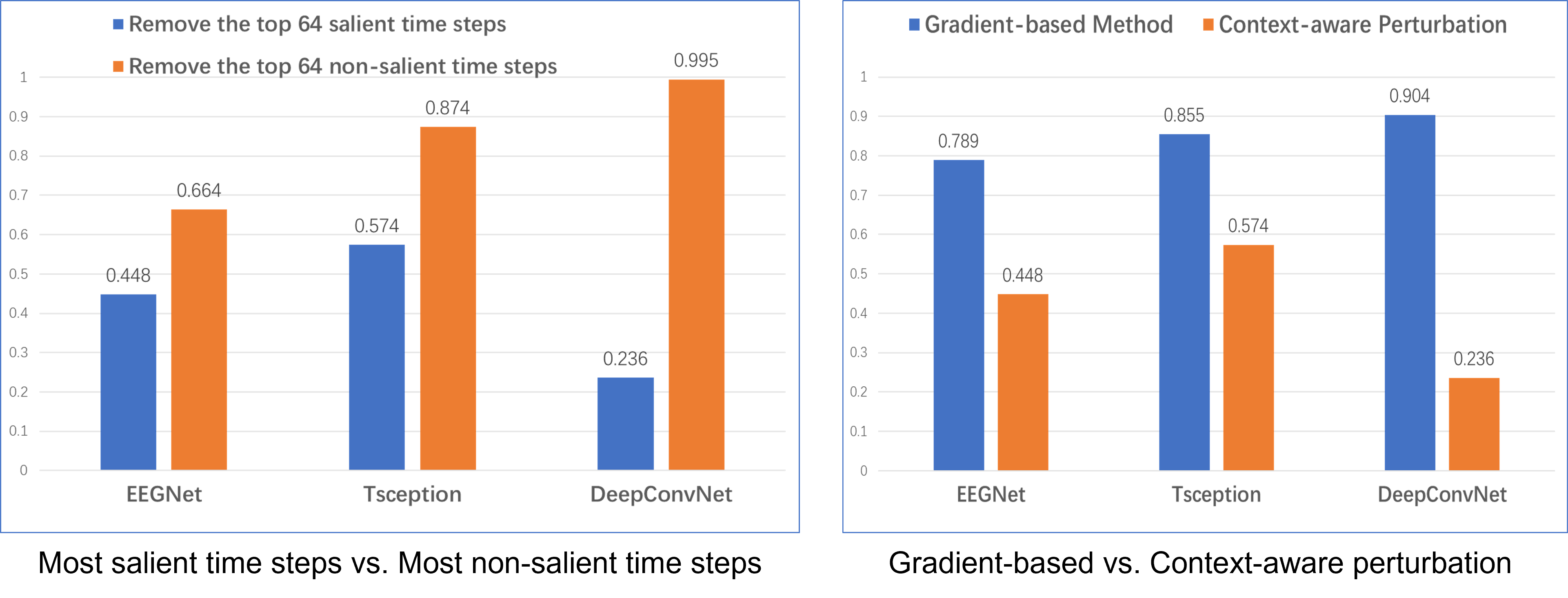}}
\caption{The comparison of mean values of accuracy. The left panel shows the comparison of removing top 64 salient time steps and removing top 64 non-salient time steps. And the right panel shows the comparison of our method with the gradient-based method.}
\label{fig6}
\end{figure}

Our method proposes an effective method to consider both temporal and spatial dependency. So, naturally, it can also identify the salient points on the temporal dimension. In this experiment, we conduct a simple instance-level experiment on the Arousal dimension of the DEAP dataset. The procedure of this experiment is very similar to the previous experiment of channel reduction. But in this experiment, we calculate the mean along the channel dimension to assess the saliency for each time step instead. To validate the identified salient time step, we remove the most salient 64 time steps and the most salient 64 time steps to see the influence on the accuracy, respectively. And we also remove the most salient 64 time steps using the saliency map generated by the gradient-based method. The result can be seen in Fig \ref{fig6}.

To begin with, compared with the removal of the top 64 non-salient time steps, the removal of the top 64 salient time steps causes a more significant decline in the mean accuracy. It proves that our method can effectively detect saliency distribution on the time dimension. Moreover, the reduced time steps identified by our method cause a more noticeable impact on the accuracy. This observation shows that our method outperforms the gradient-based method in identifying salient features on the temporal dimension. 
\subsection{Area Limitation Strategy}
\label{area}
\begin{figure}[!h]
\centerline{\includegraphics[width=\columnwidth]{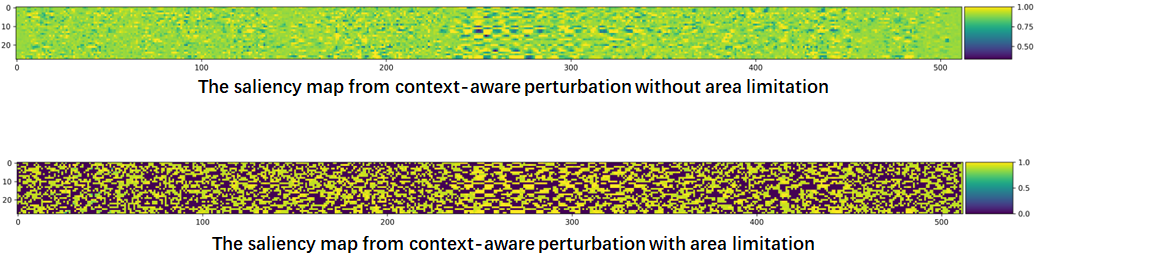}}
\caption{The comparison of saliency map with area limitation strategy. The both saliency maps are targeted at the same model and input the same sample.}
\label{fig7}
\end{figure}

In this subsection, we experiment to validate the effectiveness of our area limitation strategy. The comparison of the saliency map with area limitation strategy and the saliency map without area limitation strategy is shown in Fig \ref{fig7}. As we can see, the legibility is improved clearly. Our method turns off half of the generated saliency map, highlighting only the most informative half. Following the instance-level experiment in Section \ref{Channel Selection}, we test the effectiveness of the simplified saliency map by channel reduction. In this experiment, we remove the top k salient channel respectively, where $k\in \left\{2,4,6,8,10\right\}$. And we also compare our method with the gradient-based method.
\begin{figure}[!h]
\centerline{\includegraphics[width=\columnwidth]{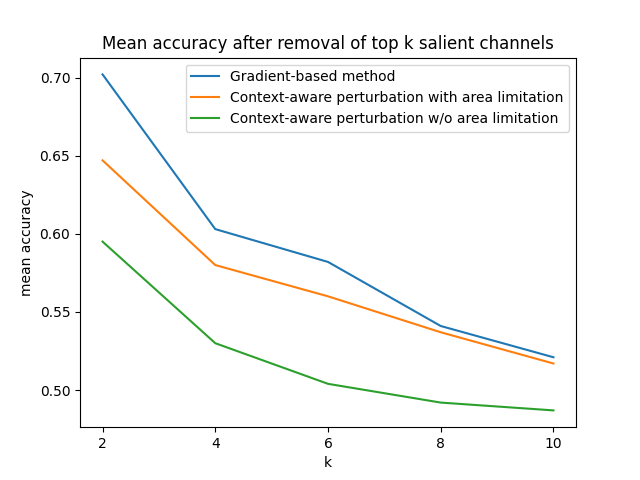}}
\caption{The mean accuracy of models when top k salient channels are removed is shown in the figure.}
\label{fig8}
\end{figure}

The results can be seen in Fig \ref{fig8}. First, we can say that our method with area limitation still outperforms the gradient-based method. This argument is based on the lower mean accuracy of our method at each k. This observation further supports the robustness of our method. Second, we admit that the effectiveness of identified saliency map suffers an obvious decline. Generally speaking, it is a trade-off between the legibility of an explanation and its effectiveness. Thus, although we adopt some measures to preserve the information of saliency, the decline in effectiveness is inevitable.
\subsection{Ablation Study}
\begin{figure}[!h]
\centerline{\includegraphics[width=\columnwidth]{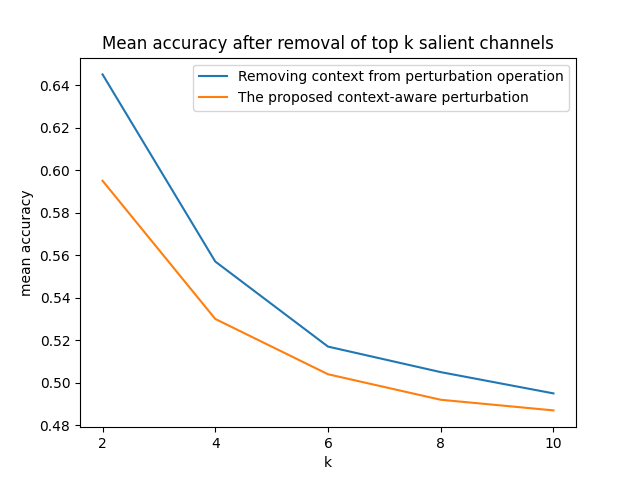}}
\caption{The result of ablation study. The impact on the model after channel reduction is shown in this figure.}
\label{fig9}
\end{figure}
The proposed context-aware perturbation incorporates the context information. To validate the effect of context on the effectiveness, we conduct an ablation study using the Arousal dimension of the DEAP dataset and TSception model. In this experiment, we remove the first term in~\eqref{eq2}, which contains the context information. And then, we implement a channel reduction experiment following the same paradigm illustrated in Section~\ref{Channel Selection}.

In Fig~\ref{fig9}, we show how the impact on the model changes according to the number of removed channels. The result shows that removing context information lessens the impact on the model. It proves that our method using context information makes a difference in the saliency identification.

\section{Conclusion}
In this paper, we review the existing works focused on explaining EEG-based deep learning models. And we find that those works are inappropriate for explaining the EEG-based model, to some extent. Considering the characteristic of EEG, we propose a context-aware perturbation method to include the temporal dependency and spatial dependency in consideration of explanation. Furthermore, we demonstrate that context-aware perturbation can also help the perturbed input get close to the distribution of normal data. It can deal with the risk of triggering artifacts in the deep learning model. We conduct extensive experiments to validate our method and make comparisons with the commonly used gradient-based method. As can be seen, our method shows a significant improvement in identifying salient elements. The results work as convincing evidence to prove that the context information is crucial for understanding the EEG-based deep learning models. 

Our experiments also reveal an interesting observation that the advance from the gradient-based method to our method might be related to the model scale. In future works, more efforts could be made to understand this phenomenon and discuss what an appropriate saliency method should be for small EEG-based models.
\bibliographystyle{IEEEtran}
\bibliography{generic-color}

\end{document}